# The Generation of Textual Entailment with NLML in an Intelligent Dialogue system for Language Learning CSIEC

Jiyou Jia

Department of Educational Technology
Peking University
Beijing, 100871, China
jjy@pku.edu.cn

## Abstract

This research report introduces the generation of textual entailment within the project CSIEC (Computer Simulation in Educational Communication), an interactive web-based human-computer dialogue system with natural language for English instruction. The generation of textual entailment (GTE) is critical to the further improvement of CSIEC project. Up to now we have found few literatures related with GTE. Simulating the process that a human being learns English as a foreign language we explore our naïve approach to tackle the GTE problem and its algorithm within the framework of CSIEC, i.e. rule annotation in NLML, pattern recognition (matching), and entailment transformation. The time and space complexity of our algorithm is tested with some entailment examples. Further works include the rules annotation based on the English textbooks and a GUI interface for normal users to edit the entailment rules.

## 1 Problem Description: Generation of Textual Entailment (GTE) and Recognition of Textual Entailment (RTE)

In daily human language we can often infer (entail) one text fragment from another one, for example "*what is the book's price?*" and "*how much is the book?*". We adopt a definition of textual entailment as a directional relationship between pairs of text expressions, denoted by T (the entailing text), and H (the entailed hypothesis). It is considered that T entails H if the meaning of H can be inferred from the meaning of T, as would typically be inferred by people (Dagan et al., 2006). This somewhat informal definition is based on (and assumes) common human understanding of language as well as common background knowledge. In some works (as Lin, et al. 2001) the entailment is called inference. That T entails H means also T infers H. So in this paper we treat these two terms as the same. This definition can be described with a formula:

T➔H, where T is the entailing text and H is the entailed hypothesis.

T➔H is true if T entails H, and this relationship is called an entailment or inference rule.

If both T➔H and H➔T are true, we call T is equivalent to H, or T⬅➔H.

The following are simple examples of text entailment and equivalence frequently used in daily life and in the system CSIEC:

(E1) *What is the price of the book?* ⬅➔ *How much is the book?*

(E2) *The student's name is Zhang.* ⬅➔ *The student is Zhang.*

(E3) *I study in Beijing University.* ⬅➔*I attend Beijing University.*

(E4) *English is his mother language.* ➔ *He can speak English.*

Two questions emerge with the text entailments:

The first is: given a text T, to calculate what can be inferred from T.

The second is: given two texts T and H, to determine whether T➔H is true or false.

We call the first question Generation of Textual Entailment (abbreviation: **GTE**), the second one Recognition of Textual Entailment (abbreviation: **RTE**). This paper attempts to solve GTE problem within the framework of CSIEC project.

## 2 The Significance of GTE to CSIEC

CSIEC (Computer Simulation in Educational Communication) is an interactive web-based human-computer dialogue system with natural lan-guage for English instruction (Jia, 2004, a). It has been put into free usage in Internet and has also been applied in English class. Despite of its achievements currently there are still some user requirements which haven't been fulfilled well, for example, the system's strong ability of natural language understating and generation, which is the fatal factor influencing the human-computer communication. The generation of textual entailment plays an important role to them.

At first the GTE is related to the redundancy of the facts database. The CSIEC can collect the user facts from the user expressions into the form of NLML, and save these facts into the table "user facts" in NLDB. The facts about the chatting robot are also treated by this way, i.e. the narrative declarative sentences about the robot are stored with the form NLML into one separate table in NLDB. But how to deal with the facts which can be inferred from other facts? If all they are stored in the database, the redundancy of the knowledge database will be greatly increased.

For example: if the user inputs "*I am an English teacher in Beijing University.*" at first interaction with the robot, this fact will be recorded in the NLDB. Later if he/she inputs "*I teach English in Beijing University.*" again, the robot should be able to recognize that the user is repeating or emphasizing a fact about him(her)self which can be inferred from the previous fact, and need not to add a new record "*I teach English in Beijing University.*" into the NLDB. Moreover, the robot should be able to generate the new sentence "*You teach English in Beijing University.*" as a response to the user's input. Although it is just a repetition of the user's input, it shows the robot's ability of logical inference, and can be regarded as a meaningful response.

Secondly the GTE is also related to the implementation of guided chatting on a given topic. For example in the guided discourse "*Salesman and customer*" if the computer knows "*what is its price?*" equals "*how much is it?*", and "*may I help you?*" equals "*can I help you?*", we need not to write all the equivalent expressions into the scenario script.

Thirdly the entailment calculation can also contribute to the question answering. In CSIEC the robot gets answers to the users' questions through scrutinizing the user facts database and the common sense knowledge database. The user facts database are enriched through the interaction between the user and the robot. If the user wants to test the intelligence of the robot, he/she may ask a question such as "*Who am I?*". Based on the user fact "*I am an English teacher in Beijing University*" the robot can answer with: "*You are an English teacher in Beijing University.*" If the user asks "W*hat do I teach?*", the answer could be readily obtained from the entailment: "*I teach English*".

At last the entailment generation can help the system's adaptation to the user language level. The vocabulary and grammar skill of a language learner varies in every learning stage. Thus the response with different levels of vocabulary and grammar skills should be generated corresponding to the learner's linguistic level. For the beginners the system should respond with a simple sentence, whereas for the advanced users the robot can speak more complicated sentence with unfamiliar words. One example is that the middle school students can understand "*This problem is difficult.*", but may not understand "*This problem is knotty.*", because they haven't learned the word "*knotty*".

So the text entailment generation is significant and critical to the evolvement of the CSIEC system. Moreover, as the RTE is associated with the GTE, we believe the solution of GTE will help the solution of RTE and other related problems such as question answering, information retrieval, etc.

# 3 Related Works to Solve GTE

Through the literature survey we can't find related works specialized on the textual entailment generation, although amounts of papers have presented the pioneer researches of discovering the inferences or checking the entailment relationship between two texts. PASCAL recognizing textual entailment (RTE) challenge (http://www.pascal-network.org/Challenges/RTE) is organized to explore what can be achieved in the area of RTE with current state-of-the-art tools. From its two past proceedings (Dagan et al. 2005; Bar-Haim et al, 2006) we can hardly find the work on GTE.

Lin and Pantel (2001) could be the first attempt to discover inference knowledge from a large corpus of text. They introduced the Extended Distributional Hypothesis, which states that paths in dependency trees have similar meanings if they tend to connect similar sets of words. Treating paths as binary relations, their unsupervised algo-rithm, DIRT (Discovering Inference from Text), can generate inference rules by searching for similar paths. A chart parser Minipar is used in DIRT to get the dependency tree. They extracted 7 million inference rules from the parse trees, among them 231,000 are unique. However, the human linguistic experts should still work to check which of the inference rules are correct one by one. The accuracy rate ranges from 0% to 92.5%.

The famous logical programming systems such as Prolog and LISP can make inferences according to the logical rules. However, the laborious transformation from natural language into exact logical language and vice versa seems to be only done by the logician experts. So in our project we attempt to directly teach the computer how to understand the textual entailment rules and how to make inference according to the entailment rules with the notation of NLML and NLOMJ, just as the English teacher teaches the students how to learn the sentence patterns and transformation rules, and how to apply them in actual language expression.

## 4 Our Naïve Approach to Solve GTE with NLML

### 4.1 Principle: rule annotation, pattern matching and entailment transformation

The entailment problem emerges with the natural language acquisition. Recall how the teachers taught us English and how we learned English as a second language. To get the entailment of example E1 in Section 1, we need just to change some words, and the other phrases remain unchanged.

Apparently only remembering this special rule is not too useful. We can describe the equivalence more generally, as the teacher taught us:
*(R1) What is the price of <something>? ⟵⟶*
*       How much is <something>?*
We call the *<something>* a pseudo variable in the rule, and "*a book*", "*those pens*", and so on, the variable's value. The entailment generation is actually the replacement of the pseudo variables in the rule right with their concrete values obtained by matching the given text with the rule left.

From the example E2 we should obtain a more generalized entailment rule:
*(R2) <person>'s name be <name>⟶ <person> be <name>.*

In this rule there are two pseudo variables. "*<person>*" represents a person, such as a student. "*<name>*" represents the concrete name, such as "*John*", "*Bill Clinton*". "*be*" represents the concrete copula form, such as "*was*", "*is*", "*has been*".

This rule doesn't contain any circumstance limitation, like time and place circumstances. This means it can be applied in any circumstance. With this generalized rule we can get the following text entailment pair:
*The student's name was John five years ago. ⟶*
*The student was John five years ago.*

We have learned these entailment rules by heart in the language course and can apply them unconsciously, though up to now we haven't discovered how our brain copes with such replacement. The rules are actually the grammatical and logical rules we learned from English textbooks. We should rewrite them into a form which the computer can understand and use. This is the first step by letting the computer learn the inference rules.

After the storage of the entailment rules the next step is generating the entailment for a given text. We can match the text with the rules in the database one by one, and find which one has the same structure as the given text. A question occurs: what does the same structure mean between two expressions? We remind again our grammatical knowledge about English. The same structure means: they both have the same mood (declarative, interrogative, imperative or exclamative), and the same sentence structure. For example the left side of the rule (R1) is a question, and its sentence structure is "subject + be + noun phrase predicate".

After finding the appropriate rule for the given text, the third step is replacing the corresponding pseudo variables of the rule right (entailment) with their concrete values, and setting the verb in the appropriate form based on the given text, mainly on the tense of the given text and the actual subject. The values can be retrieved during the matching process.

So our principle to generate textual entailment is to describe the entailment rules with the appropriate form, to search the matched rule for a given text, and then to replace the pseudo variables in the rule right with the actual values and to set the verb with the suitable form. We call the three steps rule annotation, pattern matching and entailment transformation, respectively.

## 4.2 Rule annotation: NLML with pseudo variables

NLML, as defined in (Jia, 2004, b) is a mixture of phrase tree structure and dependency tree structure, and a detailed syntactic and semantic analysis of natural language text. Almost all linguistic features (words, part-of-speech, entity type, chunk tag, grammatical function tag, head word path) are included in NLML. All kinds of grammatically right expressions, e.g. phrases and sentences, with different complexity, voice and moods, can be clearly described by NLML. Thus it can construct the basis for further syntactical and semantic analysis. For example it can be parsed into the object model of natural language expressions, NLOMJ (Jia et al., 2004), which is suitable for rule matching and rule replacement.

Limited by the paper length we just give the NLML of the example E1 left ("*what is the book's price?*"):

*<mood>question</mood>*
*<complexity>simple</complexity>*
*<subject><noun>*
*        <type>relpron</type>*
*        <word>what</word>*
*</noun></subject>*
*<verb_phrase>*
*        <verb_type>be</verb_type>*
*        <numb>sing</numb>*
*        <pers>third</pers>*
*        <tense>present</tense>*
*        <verb_word>is</verb_word>*
*  <predicate>*
*  <predicate_type>np </predicate_ type>*
*<prem><type>art</type>*
*<word>the </word></prem>*
*<noun><pers>third</pers>*
*        <numb>sing</numb>*
*        <word>price </word>*
*        <type>noun</type>*
*</noun>*
*        <prep_phrase> <prep>of</prep>*
*                <prem><type>art</type>*
*                        <word>the</word>*
*                </prem>*
*                <noun><pers> third</pers>*
*                        <word>book </word>*
*                        <numb>sing</numb>*
*                        <type>noun</type>*
*                </noun>*

We now introduce the entailment rule form and annotation with some examples. For the entailment rule (R1) left, we keep the subject and verb phrase "*what be the price*" as required pattern, change the object "*the book*" of the prepositional phrase to a pseudo variable indexed as 1, so its NLML is:

*<mood>question</mood>*
*<complexity>simple</complexity>*
*<subject><noun><type>relpron</type>*
*        <word>what</word></noun>*
*</subject>*
*<verb_phrase><verb_type>be</verb_type>*
*        <tense>present</tense>*
*        <numb>sing</numb>*
*        <pers>third</pers>*
*        <verb_word>is</verb_word>*
*  <predicate>*
*        <predicate_type>np </predicate_type>*
*        <noun><pers>third</pers>*
*        <word>price</word>*
*        <numb>sing</numb>*
*        <type>noun</type></noun>*
*        <prep_phrase>*
*            <prep>of </prep>*
*            <prem> <type>art</type>*
*                <word>the </word> </prem>*
*            <noun><pers>third </pers>*
*            <word>**pseudo variable 1** </word>*
*            <type>noun</type></noun>*
*        </prep_phrase>*
*        </predicate>*
*</verb_phrase>*

By replacing the R1 right, we should keep the predicate "*how much*" unchanged, and replace the subject with the value of "*pseudo variable 1*". But the form of "*be*" depends on the actual subject and the tense of the given text. A tag "*<verb_change/>*" is used to indicate this verb form transformation. So its NLML is:

*<mood>question</mood>*
*<complexity>simple</complexity>*
*<subject>**pseudo variable 1**</subject>*
*<verb_phrase>**<verb_change/>***
*    <verb_type>be</verb_type>*
*    <predicate>*
*    <predicate_type> query_adj </predicate_type>*

```
        <adj><adv><type> extent</type>
                 <word>how</word></adv>
              <word>much </word>
              <grad>abso</grad>
        </adj>
    </predicate>
</verb_phrase>
```

Starting from this rule we can generate the following text entailment pairs:

*What was the pen' price two years ago?* →
*How much was the pen two years ago?*

*What has the price of the bus tickets in the capital Beijing been since 2007?* → *How much have the bus tickets in the capital Beijing been since 2007?*

For the rule R2: *<person>'s name be <name>* →
*<person> be <name>*, the rule left (text) NLML is:
```
<mood>statement</mood>
<complexity>simple</complexity>
<subject>
    <noun><pers>third</pers>
     <word>name</word><numb>sing</numb>
     <type>noun</type></noun>
    <prep_phrase> <prep>of</prep>
       <prem><type>art</type>
              <word>the </word></prem>
       <noun><pers>third</pers>
              <word>pseudo variable 1</word>
       <category>person </category>
       <numb>sing</numb>
       <type>noun</type></noun>
    </prep_phrase>
</subject>
<verb_phrase><verb_type>be</verb_type>
       <numb>sing</numb>
       <pers>third</pers>
       <tense>present</tense>
       <verb_word>is</verb_word>
  <predicate>
      <predicate_type>np</predicate_type>
      <noun><word>pseudo variable 2 </word>
      <type>name</type><pers>third</pers><
      numb>sing</numb></noun>
  </predicate>
</verb_phrase>
```

Here the tag "*<category>person</category>*" indicates the noun in the prepositional phrase modifying the noun "*name*" must be an instance of the class "*person*". What is a "*person*" then? "*A student*", "*a teacher*", etc. are persons. This relationship between an occupation and a person can be retrieved from WordNet. "*Your sister*", "*his father*", etc. are persons. This family relationship can also be retrieved from the WordNet. Thus the classification of a noun phrase into the category "*person*" can be realized with WordNet. The tag "*<type>name</type>*" in the NLML has defined the predicate must be a person name, thus the new tag "*category*" is not necessary.

The entailment NLML is:
```
<mood>statement</mood>
<complexity>simple</complexity>
<subject>pseudo variable 1</subject>
<verb_phrase><verb_change>
    <verb_type>be</verb_type>
    <predicate>
        <predicate_type>np</predicate_type>
        <noun>pseudo variable 2</noun>
    </predicate>
</verb_phrase>
```
Here the tag "*<verb_change/>*" indicates the actual form of the "*be*" depends on the given text.

From example (E3) we get a more generalized rule: (R3) *Somebody studies in a given university /institute/college.* → *Somebody attends this university/institute/college.*

Its text NLML is:
```
<mood>statement</mood>
<complexity>simple</complexity>
<subject><pseudo>pseudo variable 1 </pseudo>
</subject>
<verb_phrase><voice>active</voice>
       <verb_type>verb</verb_type>
       <tense>present</tense>
       <numb>sing</numb><pers>first</pers>
       <verb_word>study</verb_word>
</verb_phrase>
<circum>
   <circum_type>prep_phrase</circum_type>
   <prep_phrase><prep>in</prep>
   <prem>
    <pers>third</pers><numb>sing</numb>
    <type>address</type><word>Beijing</word>
   </prem>
   <noun>
       <category>group</category>
```

<pers>third</pers><numb>sing</numb>
  <word>*pseudo variable 2*</word>
  <type>noun</type>
 </noun>
 </prep_phrase>
</circum>

This text pattern requires the verb part must contain the kernel verb "*study*", and there must be a prepositional phrase as the sentence circumstance, whose preposition must be "*in*" and whose noun phrase must be a kind of "*group*". Surely the "*group*" includes not only *university/institute /college,* but this rule also fits for other kinds of group.

Its entailment NLML is:
<mood>*statement*</mood>
<complexity>*simple*</complexity>
<subject>*pseudo variable 1*</subject>
<verb_phrase><**verb_change/**>
        <voice>*active*</voice>
        <verb_type>*verb_object*</verb_type>
        <verb_word>*attend*</verb_word>
        <direct_object>*pseudo variable 2*
        </direct_object>
</verb_phrase>
The entailment requires the kernel verb must be "*attend*" whose form depends on the given text, so the tag "<*verb_change/>*" is used.

From example (E4) we get a more generalized rule:
(R4) *<language noun phrase> is somebody's mother language→ somebody can speak <language noun phrase>.*

Its text NLML is:
<mood>*statement*</mood>
<complexity>*simple*</complexity>
<subject>
  <noun><pers>*third*</pers>
        <numb>*sing*</numb>
        <word>***pseudo variable 1***</word>
        <**category**>***language***</category>
        <type>*noun*</type></noun>
</subject>
<verb_phrase><verb_type>*be*</verb_type>
  <numb>*sing*</numb><pers>*third*</pers>
  <tense>*present*</tense>
  <verb_word>*is*</verb_word>
<predicate>

<predicate_type>*np*</predicate_type>
<prem><type>*possessive*</type>
  <word>***pseudo variable 2***</word></prem>
<prem><type>*noun*</type>
  <word>*mother*</word></prem>
<noun><word>*language*</word>
      <pers>*third*</pers>
      <numb>*sing*</numb><type>*noun*</type>
  </noun>
</predicate>
</verb_phrase>
This text pattern requires the subject is an instance of "*language*", and the predicate phrase is a noun phrase with a possessive pronoun as pre modifier. The possessive pronoun is a pseudo variable.

The entailment NLML is:
<mood>*statement*</mood>
<complexity>*simple*</complexity>
<subject>***pseudo variable 2***</subject>
<verb_phrase>
  <verb_type>*verb_object*</verb_type>
  <tense>*modal*</tense> <numb>*sing*</numb>
  <pers>*first*</pers>
  <verb_word>*can*</verb_word>
  <verb_word>*speak*</verb_word>
  <kernel_tense>*infi*</kernel_tense>
  <direct_object>***pseudo variable 1***
  </direct_object>
</verb_phrase>
This entailment states the verb part must be "*can speak*", so needs not the tag "<*verb_change/>*". By replacement the pronoun should be changed from its genitive (possessive) case to nominative case.

## 4.3 Pattern matching

How to check if a given text is a concrete instance of the generalized model of the rule's left? Suppose the given text is T, and the given text pattern (rule left) is P. As to all the examples mentioned above with this "simple" complexity (sentences without conjunction like "*before*", "*if*"), the algorithm of the pattern matching is:
*Compare the mood of the T with the mood of P*
*If they aren't equal, T doesn't match P*
*Else*
    *Compare the subject of T with the subject of P*
  *If the subject of T does not match the subject of P, T does not match P*
  *Else*

*Compare the verb phrase of T with that of P*
*If the former does not match the latter, T does not match P*
*Else*
*Compare the circumstances of T with that of P*

*If any circumstance in P can't find a matched circumstance in T, T does not match P*
*Otherwise T does match P*
*End*

Then more concretely, what is the matching of the given subject to the given pattern subject? As a subject is actually a noun phrase used at the sentence beginning, the comparison of two subjects is in fact the comparison of two noun phrases. By the comparison of a concrete noun phrase with the noun phrase pattern, the content of the attribute "*pseudo*" in the pattern is firstly checked. A not empty content means this pattern noun phrase doesn't have any specification; therefore the checked noun phrase matches this pattern. Afterwards the pseudo variable represented by this pattern content will be set the value of the NLML of the checked noun phrase.

An empty content of the attribute "*pseudo*" in the pattern phrase means there are specific requirements in the noun phrase. The given text should be checked more in detail with the pattern phrase.

The matching of a given verb phrase to the given pattern verb phrase requires both the verb part matching and the matching of other components. By the verb part matching at first the verb type and voice (active or passive) of the checked verb phrase will be compared with that of the pattern phrase. If either the type or voice isn't equal to that of the pattern phrase, this given verb phrase doesn't match the pattern verb phrase. Otherwise the other parts in the verb phrase will be further checked. If the verb parts of the verb phrase match that of the pattern, the other parts, e.g. predicate, or objects, should be checked further.

## 4.4 Entailment transformation

If one given text matches the left pattern of the rule, the pseudo variables in the pattern will be set the corresponding values, i.e. the NLML of the matched phrase. With them we can obtain the en-

tailment of the given text according to the rule. The algorithm is:

Replace the "*pseudo variable X*" (X is the sequence number of the pseudo variable, such as 1 and 2) in the NLML of the rule right with the corresponding actual NLML which have been obtained during the pattern matching process.

If there is tag "*verb_change*" in the entailment NLML, transfer the attributes of the verb phrase according to subject phrase and the tense of the given text.

After the verb phrase transformation we get the ultimate NLML of the entailment and then calculate the entailed text after parsing it into NLOMJ.

## 4.5 GTE Algorithm and logical entailment

Through the examples above we have showed the procedure of textual entailment generation. The algorithm for GTE can be summarized as:
*Pre process the text*
*Parse the text into NLML*
*Parse the NLML into NLOMJ*
*Compare: compare the structure of this NLOMJ with the rules in rule database*
*If there is such a rule or rules matching this one as text*
*transform the entailment NLML and calculate then the entailment texts*
*repeat Compare with the obtained entailments to get deeper entailments*
*Otherwise, employ the logical entailment algorithm to get all of its entailments.*

The logical entailment of a text is the entailment which needs not any inference rule, but can be inferred according to the logical reasoning of the common sense knowledge. For example: hypernym entailment:
*Zhang is a student.→ Zhang is the <hypernym of student>*, e.g. *Zhang is a person.*
*I have a dog→ I have a <hypernym of the dog>*, e.g. *I have an animal.*

The deeper entailment means the entailment of the entailment. To avoid repeating reversed entailment, we label each rule with an identification number, and specify the number of its reversed rule. For example if we label the rule R3 with the ID=3, and label its reversed rule R5 with ID=5, the R3's reversed rule ID is 5, and vice versa.

(R5) *somebody attends this university/institute /college.* → *Somebody studies in a given university /institute/college.*

### 4.6 Assistant authoring tool to edit the entailment rules: TEE

Apparently it is too difficult for a normal English teacher to edit the entailment rules. We have designed a Java GUI, a so called TEE (Textual Entailment Editor), to assist the normal user to edit the rule easily. The rule annotator needs not to remember the NLML in details, but only input one example pair of text and entailment. The TEE then will interactively guide the annotator to make some choices by just clicking on buttons, and finally get the entailment rule. At the end the annotator can check the rule with new texts. Of course the annotator should be good at English grammar.

## 5    Implementation and Complexity Prediction

We are cooperating with English teachers to manually build the entailment rule database for the textual entailment occurring in the textbooks of schools and universities with TEE. An annotator can write 10 rules in one hour. This method is laborious; however, it is a reliable one. If it is impossible for the human being to write all of the rules implicit hidden in the giant mounts of corpus, it is still plausible to write the rules taught in the English courses from elementary school to university, which can contribute much to improve the intelligence of CSIEC system. The entailments rules and their test can be accessed in the CSIEC website.

Besides the rules which can be easily retrieved from the English textbooks, we are also planning to use the resources available from Internet, for example, Sekine's Paraphrase Database (http://nlp.cs.nyu.edu/paraphrase/), and TEASE (http://aclweb.org/aclwiki/index.php?title=TEASE). The former includes a paraphrase database by Hasegawa's method (Hasegawa et al. 2004) with 755 sets of paraphrases, and 3,865 phrases in total, which have been cleaned up by human annotator, as well as the paraphrase database by Sekine's method (Sekine 2005) with 19,975 sets of paraphrases, and 191,572 phrases in total, which have not been cleaned up by human. The latter (Szpektor et al.

2004) consists of 136 different templates, every of which is a set of entailment relations.

To predict the complexity of the GTE algorithm dealing with more and more entailment rules, we make a test with 10,000 rules, among them 20 are unique and the others are the same, what will not reduce the comparison time. The generation of textual entailment for a given text costs 100ms or so time, almost the same as the testing with just 20 unique rules. But the memory cost is linearly proportional to the rules number. The 10,000 rules occupy about 300 Megabytes physical memory.

## 6    Discussion

The underlying idea of our GTE algorithm is very naïve: the language teachers told us the sentence pattern and inference rules, we learn by heart these rules one by one and apply them in thinking and speaking. No almighty method can be given by the teacher and learned by us so that we can use it to all entailment generation. In language education, this is an old, traditional and plausible way. But to the best of our knowledge, no researcher in computer science and NLP has used it to model the GTE in computer. Maybe the algorithm complexity by dealing with the seemingly very large amount of rules is the main obstacle (Stefik, 1995), .e.g. the 231,000 unique inference rules found in (Lin and Pantel, 2001). We will also face this problem, as the rule database grows.

Additionally, the rule annotation with NLML and its machine interpretation will become more complicated. In this paper we just illustrated the GTE algorithm with the example of simple sentences with only one subject plus one verb phrase structure. More work should be done to solve the GTE of complex sentences with subordinate sentences.

We begin the research of GTE from our interactive language learning project CSIEC with the objectives to reduce the facts redundancy, to generate reasonable and diverse responses, and so on. So the evaluation of the application results should be implemented in the future. Moreover, we will attempt to use the GTE approach to tackle other hard problems in NLP, such as RTE, question answering and information retrieval, etc.

## Acknowledgments

We thank the support to our projects from Ministry of Education China, Beijing University, and Education Committee of Capital Beijing.